\documentclass{article}

\usepackage[final,nonatbib]{nips_2017}

\usepackage[utf8]{inputenc} 
\usepackage[T1]{fontenc}    
\usepackage{hyperref}       
\usepackage{url}            
\usepackage{booktabs}       
\usepackage{amsfonts}       
\usepackage{nicefrac}       
\usepackage{microtype}      
\usepackage{amsmath}
\usepackage{amssymb}
\usepackage{amsthm}
\usepackage{graphicx}
\usepackage{color}

\usepackage{biblatex}
\bibliography{main}
 
\theoremstyle{definition}
\newtheorem{definition}{Definition}[section]
\newtheorem{statement}{Statement}[section]

\title{Feature importance scores and lossless feature pruning using Banzhaf power indices}
\author{Bogdan Kulynych\thanks{\texttt{hello@bogdankulynych.me}} \qquad Carmela Troncoso\thanks{\texttt{carmela.troncoso@epfl.ch}} \\
École Polytechnique Fédérale de Lausanne (EPFL) \\
}
\date{}

\begin{document}

\maketitle

\begin{abstract}
Understanding the influence of features in machine learning is crucial to interpreting models and selecting the best features for classification. In this work we propose the use of principles from coalitional game theory to reason about importance of features. In particular, we propose the use of the Banzhaf power index as a measure of influence of features on the outcome of a classifier. 
We show that features having Banzhaf power index of zero can be losslessly pruned without damage to classifier accuracy. Computing the power indices does not require having access to data samples. However, if samples are available, the indices can be empirically estimated. We compute Banzhaf power indices for a neural network classifier on real-life data, and compare the results with gradient-based feature saliency, and coefficients of a logistic regression model with $L_1$ regularization.
\end{abstract}

\section{Introduction}
Evaluating importance of features allows to better understand the behaviour of trained models, and can be used to select features for improving models' performance and efficiency.
Consider a classifier $F: \mathcal{X} \rightarrow \{0, 1\}$ that operates in a binary feature domain $\mathcal{X} = \{0, 1\}^n$. For such models, we propose the use of Banzhaf power indices from coalitional game theory as a measure of feature importance. We show how this index can be used as a feature importance score, and how it can be used for lossless feature pruning. Additionally, we argue that Banzhaf power index may be useful in evaluating the adversarial robustness of models.

We introduce the basic theory behind Banzhaf power indices, and refer to existing work for techniques for computing the indices. We carry out an experiment on a real dataset, comparing the indices as feature importance scores with other well known approaches: gradient-based saliency (e.g., \cite{steppe1997feature}), and coefficients of a logistic regression model trained with $L_1$ weight regularization \cite{ng2004feature}. We finally discuss the limitations of this approach and point for future research directions.

\section{Theoretical background}

A simple coalitional game is a pair $(\mathcal{A}, F)$, where $\mathcal{A}$ is a finite set of players, and $F$ is a mapping $2^A \rightarrow \{0, 1\}$ from coalitions of players to game outcomes.

To look at machine learned classifiers as simple coalitional games, we first fix \emph{feature space} $\mathcal{X}$ as a set of all possible binary vectors $\{0, 1\}^n$. Players in a game correspond to classifier features, and coalitions of players correspond to examples, or feature vectors, from $\mathcal{X}$. With a slight abuse of notation, we use coalitions and their respective binary indicator representations interchangeably. Having that, we can view a trained classifier $F: \mathcal{X} \rightarrow \{0, 1\}$ as a simple coalitional game $(\mathcal{A}, F)$, where $\mathcal{A}$ are the classifier features, with $|\mathcal{A}| = n$.

Even though it may seem that most real-world problems machine learning use more complex feature sets, it is relatively easy to transform many datasets into this binary form. For example, continuous features can be quantized, and categorical features can be one-hot encoded. 

To apply the framework from coalitional game theory to computing feature importance scores, we introduce the notions of marginal contribution and the Banzhaf power index.

\begin{definition}
\emph{Marginal contribution} of a player $i \in \mathcal{A}$ to a coalition $S \in 2^\mathcal{A}$ is a function:
$$\Delta_i F(S) = F(S \cup \{i\}) - F(S)$$
\end{definition}

\begin{definition}
A player $i$ is \emph{critical} to a coalition $S$ if $$|\Delta_i F(S)| = 1$$
\end{definition}

It is common in literature that games with non-decreasing $F$ are discussed. If that is the case, the condition for player being critical is defined as $\Delta_i F(S) = 1$. This means that a player $i$ is critical if they make a \emph{winning coalition} (that is, a coalition $S \cup \{i\}$, such that $F(S \cup \{i\}) = 1$) a \emph{losing} one (respectively, $F(S) = 0$) when leaving. The games that correspond to most common classifiers in machine learning are not necessarily non-decreasing. Hence, the definition above additionally considers a player critical if they make a losing coalition a winning one when leaving.

The Banzhaf power index, initially introduced in \cite{banzhaf1964weighted}, is defined as the expected number of coalitions for which a player is critical, if all coalitions were equally likely.

\begin{definition}
\emph{Banzhaf power index} of a player $i$:

$$I_f(i) = \frac{1}{2^{n - 1}} \sum_{S \in 2^\mathcal{A}} |\Delta_i F(S)| = 1$$
\end{definition}

Equivalently, we can write:

$$I_f(i) = \mathbb{E}_{S \sim U[2^\mathcal{A}]} |\Delta_i F(S)| = 1, $$

where $U[2^\mathcal{A}]$ is the uniform distribution over all coalitions. 

A straightforward generalization is the \emph{weighted Banzhaf power index}, which extends the index to a possibly non-uniform distribution $p(x)$ over coalitions:

$$I_{f, p}(i) = \mathbb{E}_{S \sim p(x)} |\Delta_i F(S)| = 1.$$

We also introduce the notion of \emph{empirical Banzhaf power index}:

$$I_{f, X}(i) = \frac{1}{2^{|X|}}\sum_{S \in X} |\Delta_i F(S)| = 1,$$

for some $X \subset 2^\mathcal{A}$.

Finally, some notions that are useful for feature pruning.

\begin{definition}
A player $i$ is \emph{dummy} if there exist no coalitions $S \in 2^\mathcal{A}$, for which the player is critical.
\end{definition}

It is easy to see that a player is dummy if and only if for all $S \in 2^\mathcal{A}$ it holds that $\Delta_i F(S) = 0$.

It is also trivial to show the following result:

\begin{statement}
A dummy feature $i$ has Banzhaf power index $I_f(i) = 0$.
\end{statement}

\section{Applications of Banzhaf power indices to machine learning}

The power index $I_{f}(i)$ shows how often the feature $i$ can flip a classification outcome, with all possible feature vectors considered equally likely. In practical machine learning tasks, feature vectors do not tend to be distributed with equal probabilities over $\{0, 1\}^n$. The empirical power index variant $I_{f, X}$ can show the ``ability'' of a feature to flip classifications on a data sample, which should reflect the real data distribution.
We can therefore use it as one possible measure of feature importance.

\textbf{Lossless pruning.} The regular Banzhaf power index, however, can be useful in identifying dummy features. Such features can not possibly flip the classification outcome: $F(S \cup \{i\}) = F(S)$ for any $S$. Let $F(S)$ be computed by thresholding the output of a score $f(S)$ (confidence). If $i$ is a dummy feature, then the scores $f(S)$ and $f(S \cup \{i\})$ may be different even though the classification outcome after thresholding must be the same. If the model $F$ is used in decision-only mode, that is, the confidence is not used in an application, then excluding such dummy features is lossless: classification outcomes do not change for \emph{any possible feature vector}. Therefore, pruning features having a Banzhaf power index of zero does not hurt accuracy.

\textbf{Adversarial robustness.} Additionally, the fact that Banzhaf power index considers all possible feature vectors equally may be an advantage in adversarial settings. A commonly considered scenario (see, for example, \cite{goodfellow2014explaining}) is producing an \emph{adversarial example} $\hat S$. The adversary attempts to find $\hat S$ based on a real example $S$, such that $F(\hat S) \neq F(S)$, while keeping $\hat S$ and $S$ ``close'' in some sense. Even though the examples have to be ``close'', the adversary does not need to be constrained by the original data distribution. In this context, Banzhaf power index is useful, since it shows the feature importance over all possible feature vectors.

Features with low Banzhaf power index are not useful to an adversary that needs to flip the classification outcome, and vice verca, features with high power index are a good target for manipulation.

\subsection{Computation}

There exist efficient algorithms for computing the power index using generating functions \cite{bilbao2000generating} in the classical case of weighted voting games with non-negative integer weights. In our analogy this corresponds to linear models with non-negative integer weights, which is a quite limited class of models. Generally, the computation is exponential in the number of features.

To practically approximate the Banzhaf power index $\hat I_f(i)$ for models with small number of features, a probabilistic approximation approach \cite{bachrach2008approximating} can be used. For that, one needs to sample $k$ random coalitions $S \leftarrow U[2^\mathcal{A}]$, where $k$ can be determined from a desired approximation accuracy $\varepsilon$ and probability $1 - \delta$ of the true value lying within $(\hat I_f(i) - \varepsilon, \hat I_f(i) + \varepsilon)$ confidence interval. This solution, unfortunately, suffers from the curse of dimensionality: the larger $n$ gets, the smaller $\varepsilon$ needs to be to produce meaningful estimates.

\section{Evaluation}

To see how the results of feature importance scoring using Banzhaf power indices compare to other methods, we train a neural network on SPECT dataset~\cite{lichman2013}. This is a small real-life dataset  comprising 267 cardiac single proton emission computed tomography (SPECT) image representations, each having 22 binary features. We choose this dataset because of the practical constraints for computing the exact Banzhaf power indices (22 features is small enough to compute in reasonable time).

We use a single hidden layer neural network with 20 neurons and ReLU activations. For a trained network $f$, we compute average gradient-based saliency as follows:

\begin{equation}\label{eq:saliency}
I'_f(i) = \sum_{x \in X} |\nabla_i f(x)|
\end{equation}

where $X$ is the full data set. Additionally, we train a logistic regression with cross-entropy loss and $L_1$ weight regularization, and use the values of obtained coefficients as a measure of importance.

For both models, we split the dataset into training and testing subsets of 187 and 80 examples respectively. The neural network has attained 86\% accuracy on the test set, and the logistic regression model has attained 72\%.

Finally, we compute the exact Banzhaf power indices, and empirical indices on the full dataset for the neural network.

\begin{figure}
    \centering
    \includegraphics[width=\textwidth]{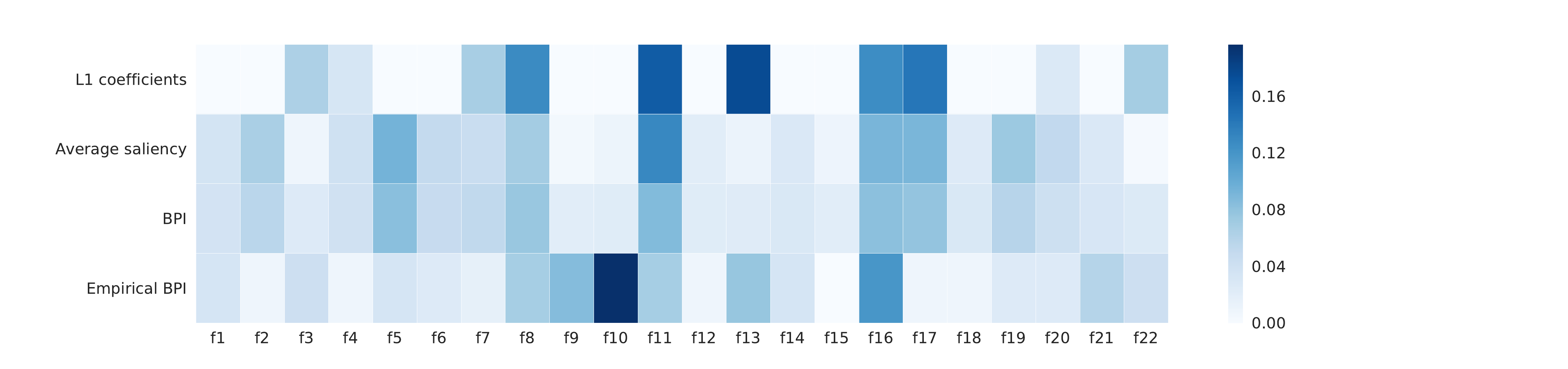}
    \caption{Comparison of feature importance scores using different methods: \emph{\footnotesize{(1)~coefficients of a logistic regression model trained with $L_1$ regularization, (2) average gradient magnitude w.r.t. inputs over the dataset, (3) exact Banzhaf power index, and (4) empirical Banzhaf power index on the dataset}}}
    \label{fig:spect_plot}
\end{figure}

\textbf{Results.} We show the obtained values for each feature (\texttt{f1}--\texttt{f22}) in Fig.~\ref{fig:spect_plot}. There is some expected overlap between high-scoring features across all techniques (e.g. \texttt{f11}, \texttt{f16}), and low-scoring ones (e.g., \texttt{f9}, \texttt{f19}). However, empirical Banzhaf indices highlight features that were not given high importance scores by other methods (e.g., \texttt{f10} was pruned in $L_1$ coefficients).

The important insight from this experiment is that the scores obtained by averaging the gradients in Eq.~\ref{eq:saliency} is very similar to the exact Banzhaf power index. We discuss this in detail in the following section.

\section{Conclusion and future research lines}

In this work we have discussed that Banzhaf power indices can be a useful tool to measure the importance of features in machine learning classification, allow for lossless feature pruning, and moreover, can be used for evaluation of adversarial robustness.

In practice, however, the power indices have to be approximated, or computed with exponential complexity by traversing all possible feature vectors. This is not practical for anything but models working on a small number of features. It is not clear whether it is possible to find efficient ways to compute the indices for anything other than weighted voting games with non-negative integer weights. If this was possible, we see lossless feature pruning as an important application, especially for large models. For example, expanding the class of games for which the computation is efficient to weighted games with non-integer weights would allow to apply the pruning to a much wider class of linear models.



In our experiment the Banzhaf power indices almost coincided with averaged values of model gradients with respect to inputs over the dataset. The opposite fact, that weights in voting games (which are the gradient of a corresponding linear model) sometimes do not correspond to actual voter power was the motivation behind introducing the power indices in the first place.

\printbibliography

\end{document}